\documentclass[conference]{IEEEtran}
\IEEEoverridecommandlockouts
\usepackage{cite}
\usepackage{amsmath,amssymb,amsfonts}
\usepackage{algorithmic}
\usepackage{graphicx}
\usepackage{textcomp}
\usepackage{xcolor}
\usepackage{hyperref}
\usepackage{tabularx}
\usepackage{booktabs}
\usepackage{multirow}
\usepackage{float}

\def\BibTeX{{\rm B\kern-.05em{\sc i\kern-.025em b}\kern-.08em
    T\kern-.1667em\lower.7ex\hbox{E}\kern-.125emX}}
\begin{document}

\title{ResNetVLLM-2: Addressing ResNetVLLM's Multi-Modal Hallucinations
 \\
\thanks{University of Windsor.}
}
 
\author{\IEEEauthorblockN{Ahmad Khalil}
\IEEEauthorblockA{\textit{Dept of Computer Science} \\
\textit{University of Windsor}\\
Windsor, Canada \\
https://orcid.org/0009-0009-4483-5839}
\and
\IEEEauthorblockN{Mahmoud Khalil}
\IEEEauthorblockA{\textit{Dept of Computer Science} \\
\textit{University of Windsor}\\
Windsor, Canada \\
https://orcid.org/0009-0001-4112-0848}
\and
\IEEEauthorblockN{Alioune Ngom}
\IEEEauthorblockA{\textit{Dept of Computer Science} \\
\textit{University of Windsor}\\
Windsor, Canada \\
angom@uwindsor.ca}
}

\maketitle

\begin{abstract}
Large Language Models (LLMs) have transformed natural language processing (NLP) tasks, but they suffer from hallucination, generating plausible yet factually incorrect content. This issue extends to Video-Language Models (VideoLLMs), where textual descriptions may inaccurately represent visual content, resulting in multi-modal hallucinations. In this paper, we address hallucination in ResNetVLLM, a video-language model combining ResNet visual encoders with LLMs. We introduce a two-step protocol: (1) a faithfulness detection strategy that uses a modified Lynx model to assess semantic alignment between generated captions and ground-truth video references, and (2) a hallucination mitigation strategy using Retrieval-Augmented Generation (RAG) with an ad-hoc knowledge base dynamically constructed during inference. Our enhanced model, ResNetVLLM-2, reduces multi-modal hallucinations by cross-verifying generated content against external knowledge, improving factual consistency. Evaluation on the ActivityNet-QA benchmark demonstrates a substantial accuracy increase from 54.8\% to 65.3\%, highlighting the effectiveness of our hallucination detection and mitigation strategies in enhancing video-language model reliability.
\end{abstract}


\section{Introduction}
\IEEEPARstart{L}{arge} Language Models (LLMs) have revolutionized natural language processing (NLP) tasks, enabling applications such as content generation, information retrieval, and conversational agents. However, despite their impressive capabilities, LLMs are prone to a phenomenon known as hallucination, where they generate content that sounds plausible, but is factually incorrect or non-existent \cite{Maynez2020Faithfulness}. Hallucinations can manifest as fabricated facts, incorrect citations, or entirely fictional entities \cite{Ji2022Survey}, posing challenges to the reliability and trustworthiness of these models, particularly in high-stakes domains such as healthcare, law, and finance.

LLM hallucination can be categorized into two main types: intrinsic and extrinsic \cite{Huang_2025}. Intrinsic hallucination occurs when the generated content conflicts directly with the provided source context. For example, when summarizing a news article, an LLM may introduce details that are not present in the original text. Extrinsic hallucination, on the other hand, refers to content that cannot be verified against the given context or external knowledge, often introducing ungrounded or fabricated information.

The causes of LLM hallucination stem from the fundamental nature of how these models are trained. LLMs predict the next token based on statistical patterns in massive datasets, rather than reasoning through factual correctness. Consequently, incomplete, biased, or misleading prompts can lead to fabricated outputs. Furthermore, LLMs lack fact-checking mechanisms and are trained on large, unlabeled data sets, making it impractical to ensure factual accuracy across all generated content \cite{Huang_2025}.

Recently, multi-modal vision-language models have introduced a new dimension to the hallucination problem \cite{Tong2024EyesWideShut, Guan2023Hallusionbench, Liu2023VisualInstruction}. These models, which combine visual and textual modalities, suffer from multi-modal hallucination, where the generated textual descriptions do not accurately reflect the visual content. This misalignment can lead to entirely fictional or misleading interpretations of visual data. 

VideoLLMs inherit hallucinations from their base LLMs \cite{Tong2024EyesWideShut}. Since these models are often built on top of pretrained LLMs, they are prone to generating fabricated content, even when interpreting visual data \cite{Tong2024EyesWideShut}. Furthermore, VideoLLMs frequently over-rely on textual priors, causing them to "guess" missing visual details based on prior language associations rather than actual visual evidence \cite{Liu2023VisualInstruction}. This can lead to inaccurate interpretations. For example, a video-generated caption describes a second railway track on the Glenfinnan Viaduct in Scotland (as shown in Fig.~\ref{fig:example-vllm-hallucination-Glenfinnan}), despite the viaduct having only one track in reality (as shown in Fig.~\ref{fig:real_glenfinnan_viaduct}).

Given the growing reliance on VideoLLMs in real-world applications such as video analysis, autonomous systems, and multimedia search, addressing hallucination is critical. Our work focuses on ResNetVLLM \cite{khalil2024resnetvllm}, a vision-language model that suffers from multi-modal hallucination. To detect hallucination in ResNetVLLM, we use a faithfulness detection strategy, where an LLM acts as a judge to evaluate the accuracy of the generated content. Additionally, we employ a modified version of the Lynx model \cite{ravi2024lynxopensourcehallucination}, originally designed for text, adapted for video by using the ActivityNet Captions dataset \cite{krishna2017dense} instead of HaluBench \cite{ravi2024lynxopensourcehallucination}. This model assesses the faithfulness of the output by comparing it against reference data, detecting unsupported or contradictory information. To mitigate this issue, we explore Retrieval-Augmented Generation (RAG) \cite{Huang_2025}, which integrates external knowledge to enhance response accuracy. As part of RAG, we introduce a modified version of Post-Hoc Retrieval \cite{Huang_2025}, where we dynamically build an ad-hoc knowledge base during inference to cross-verify and enrich the generated content. This approach ensures that the information retrieved is more reliable and contextually accurate.

The main contribution of the paper is the introduction of the second version of ResNetVLLM, which incorporates novel techniques to mitigate multi-modal hallucinations in VideoLLMs. This improves the performance of ResNetVLLM on the ActivityNet-QA benchmark, increasing its accuracy from 54.8\% to 65.3\%.

\section{Related Work}
\label{sec:related_work}

\textbf{ResNetVLLM: A Video-Language Model.} ~ResNetVLLM \cite{khalil2024resnetvllm} is a state-of-the-art model designed for generating textual responses from video inputs. It integrates a Residual Neural Network (ResNet) visual encoder with a large language model (LLM) to enhance video understanding and generate descriptive captions. The architecture consists of a ResNet module for visual feature extraction, followed by a language model for generating coherent textual responses. This combination allows ResNetVLLM to effectively interpret complex video content and produce contextually accurate captions.

Figure~\ref{fig:resnetvllm-architecture} illustrates the overall architecture of ResNetVLLM, showing the interplay between the ResNet encoder and the LLM. Additionally, Figure~\ref{fig:resnetvllm-output} demonstrates a sample output from the model, highlighting its ability to generate detailed captions. Table~\ref{tab:zero-shot-evaluation} shows the results of ResNetVLLM in the Zero-Shot Question-Answer Evaluation on the ActivityNet-QA dataset \cite{yu2019activitynet}. The model outperforms recent state-of-the-art methods, achieving a score of 54.8\%, which is higher than competing models such as Valley (45.1\%) and BT-Adapter (45.7\%).

\textbf{A Survey on Hallucination in Large Language Models: Principles, Taxonomy, Challenges, and Open Questions.} ~~Hallucination detection has gained attention in the context of large language models (LLMs). Huang et al. \cite{Huang_2025} conducted a comprehensive survey on hallucinations in LLMs, categorizing various types and exploring detection methodologies. Figure~\ref{fig:hallucination-survey} outlines the main content flow of their survey, including hallucination causes, detection strategies and evaluation benchmarks, as well as mitigation techniques.

\textbf{Lynx: Detecting Hallucinations in LLMs.} ~~Lynx, developed by Ravi et al. \cite{ravi2024lynxopensourcehallucination}, is an open-source model designed to assess hallucinations in LLM outputs. It evaluates the faithfulness of generated responses by comparing them to a reference context. The model uses the HaluBench dataset \cite{ravi2024lynxopensourcehallucination}, which contains 15K real-world samples, for evaluation. Lynx detects unsupported or contradictory information by measuring the semantic similarity between the LLM-generated text and the reference data.  The model demonstrates high accuracy in identifying hallucinated content, making it a valuable tool for assessing the reliability of video-language models such as ResNetVLLM.

\section{Method}

\subsection{Problem Definition}
Given a video \( v \) and a video captioning model \( P \), the model is said to be hallucinating if the generated caption \( P(v) \) lacks consistency with the ground-truth context \( C(v) \). In this context, \( P(v) \) refers to the \textit{generated caption} produced by the model \( P \) for a given video \( v \), while \( C(v) \) refers to the \textit{ground-truth context} or \textit{reference context} associated with the video. This context contains the factual and accurate description of the video, serving as the basis for evaluating the correctness of the generated caption.

\vspace{0.25cm}
Formally, the hallucination condition can be expressed as:
\[
    \text{Hallucination} = 
    \begin{cases} 
    \text{True}, & \text{if } P(v) \not\subseteq C(v) \\
    \text{False}, & \text{otherwise}.
    \end{cases}
\]

This approach ensures that any information in the generated output that lacks grounding in the reference context is flagged as hallucination.

\subsection{Model Formulation}
Here, we discuss our protocols for hallucination detection and mitigation in ResNetVLLM.

\textbf{Hallucination Detection Protocol.} ~~Our hallucination detection protocol identifies and measures hallucinations in ResNetVLLM by evaluating the semantic consistency between the generated captions and the corresponding video content. Since ResNetVLLM generates textual descriptions from video frames and metadata, ensuring alignment between the generated captions and the visual content is essential for reliable hallucination detection.

To detect hallucination, we adopt the \textit{Faithfulness Detection Strategy}, which assesses the factual consistency between the generated captions and the video content. This strategy combines two complementary approaches: \textit{Fact-based Detection}, which measures the overlap of key facts between the generated caption and the reference video content to identify missing or contradictory information; and \textit{LLM-based Detection}, which leverages large language models as judges by using evaluation prompts to determine whether the generated caption is semantically faithful to the video context.

Our hallucination detection model is a modified version of the \textit{Lynx hallucination detection model}, which was originally designed to detect hallucinations in text-to-text models. We extend Lynx to handle video-to-text models by replacing the text input with video data—specifically, we use the ActivityNet Captions dataset \cite{krishna2017dense} instead of the HaluBench dataset, allowing the model to evaluate hallucinations in video-generated captions. Additionally, we fine-tune the modified Lynx model for video content, enabling it to take pairs of generated and ground-truth video captions as input and assess their consistency.

\begin{figure}
    \centering
    \includegraphics[width=0.95\linewidth]{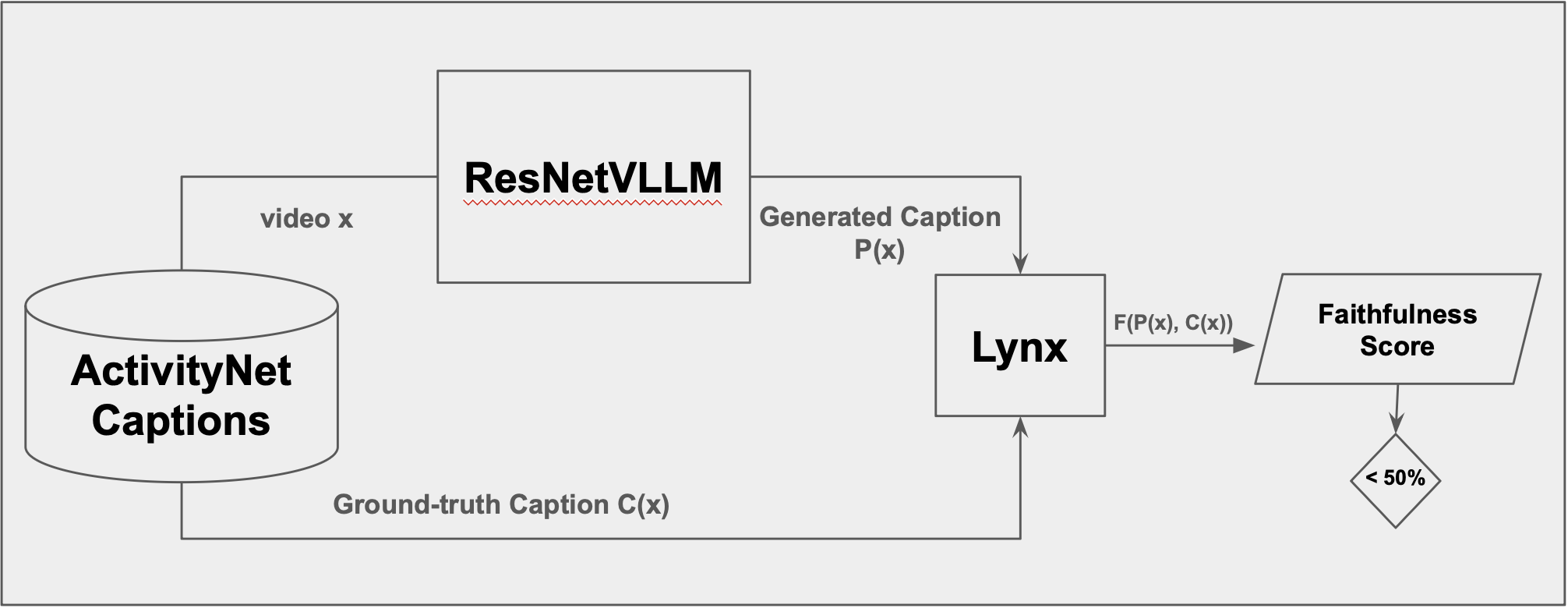}
    \caption{Pipeline of the hallucination detection protocol in ResNetVLLM}
    \label{fig:hallucination_detection_model}
\end{figure}

The \textit{hallucination detection protocol} involves a multi-step process. First, the ResNetVLLM model takes a video input \( x \) from the ActivityNet Captions dataset and generates a corresponding caption \( P(x) \). The dataset provides ground-truth captions \( C(x) \) for 20K untrimmed videos, which serve as the reference context. Next, the modified Lynx model takes the generated caption \( P(x) \) and the ground-truth caption \( C(x) \) as input and computes the faithfulness score \( F(P(x), C(x)) \) for each pair. Finally, hallucinations are detected based on a threshold: faithfulness scores below 50\% indicate hallucination. The overall faithfulness score is calculated as the average across all pairs, as shown in Equation~\ref{equ:faithfulnness_score}. The complete hallucination detection pipeline is illustrated in Figure~\ref{fig:hallucination_detection_model}.

\textbf{Hallucination Mitigation Protocol.} ~~To reduce hallucinations in ResNetVLLM, we introduce our Hallucination Mitigation Protocol, which incorporates a \textit{Retrieval-Augmented Generation (RAG)} mechanism and an \textit{Ad-hoc Knowledge Base}. RAG is a hybrid approach that combines retrieval and revision of generated results. It integrates external knowledge retrieval with the model’s generation process, improving the factual accuracy of the outputs. During inference, RAG performs a semantic search on an external knowledge base, retrieving relevant context. The retrieved knowledge is then fused with the video frames and metadata, conditioning the caption generation process. This additional contextual information reduces hallucination by anchoring the generated captions in factual evidence.

We leverage RAG to fact-check the generated output by verifying it against a grounded knowledge base. However, rather than relying on a static external source, we introduce a \textit{modified RAG approach} by dynamically building an \textit{ad-hoc knowledge base} during inference, specifically tailored to the input video content. This knowledge base is populated with the extracted projection layer outputs from ResNetVLLM, ensuring access to the most relevant and contextual information during generation. By grounding the captions in reliable, factual evidence, this strategy effectively reduces hallucinations.

The general idea behind our mitigation protocol is to create an ad-hoc knowledge base from ResNetVLLM’s projection layer outputs and integrate a Post-Hoc RAG module at the final step. The RAG module reviews and verifies the model’s output against the retrieved knowledge, ensuring factual consistency. 

The \textit{hallucination mitigation protocol} involves a series of steps designed to enhance the factual accuracy of generated captions. The input video is first fed into the ResNet module of ResNetVLLM to extract visual features. These features, along with relevant video metadata, are stored in an \textit{ad-hoc knowledge base} that is dynamically tailored to the input video content. The extracted visual features are then passed to the LLM module of ResNetVLLM to generate the corresponding video captions. To verify the generated output, a \textit{Post-Hoc RAG} module is integrated at the final stage of ResNetVLLM, retrieving supporting evidence from the knowledge base. The generated caption is then checked to determine whether it is grounded in the retrieved evidence. If discrepancies are identified between the caption and the supporting information, the response is revised to enhance its factual precision. 

\begin{figure}
    \centering
    \includegraphics[width=0.75\linewidth]{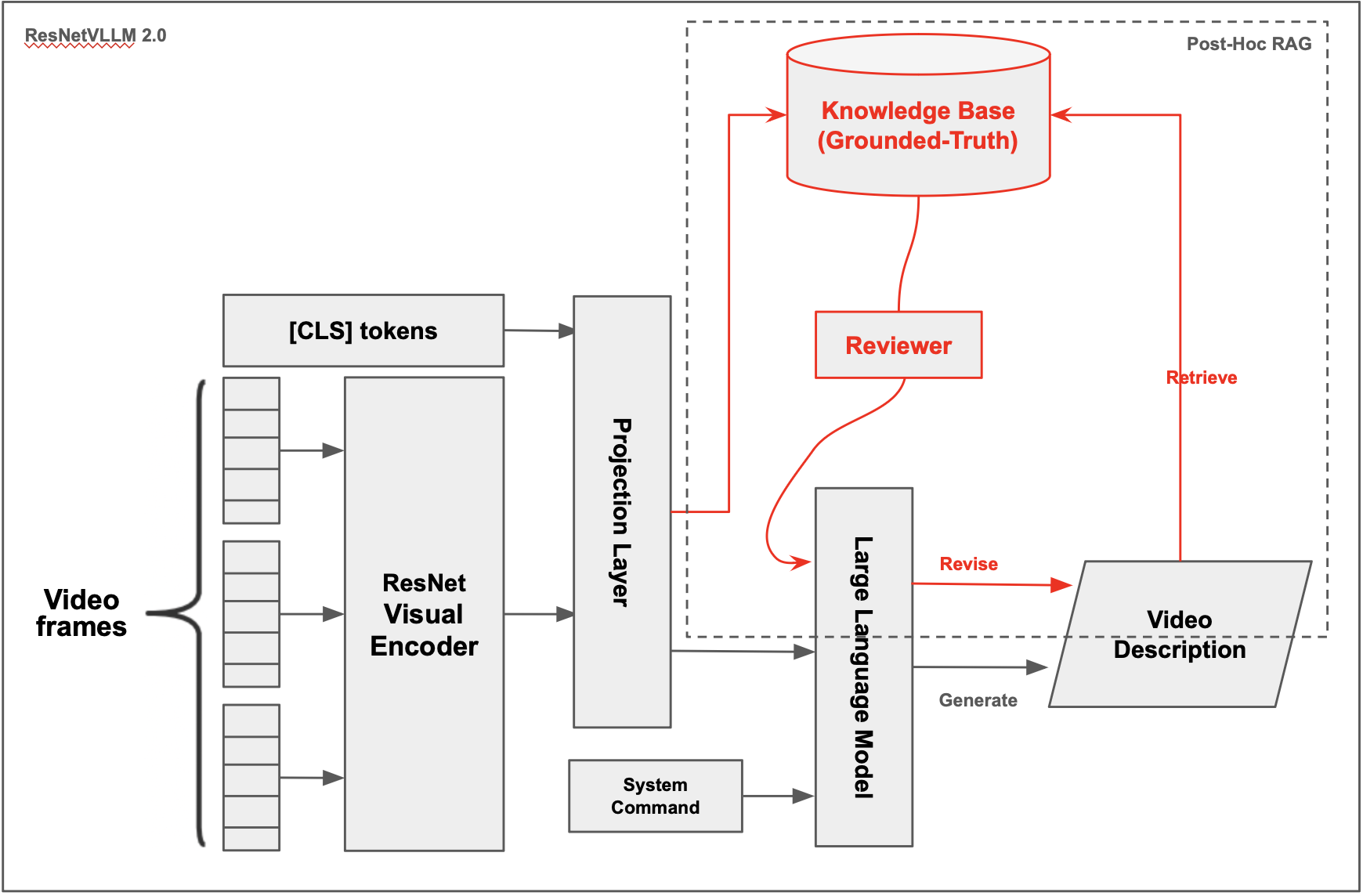}
    \caption{Overall architecture of the ResNetVLLM-2 model, including the hallucination mitigation module.}
    \label{fig:resnetvllm2-architecture}
\end{figure}

This protocol improves the performance of the model, resulting in the second version: \textbf{ResNetVLLM-2}, as illustrated in Figure~\ref{fig:resnetvllm2-architecture}.

\section{Experimental Evaluation}
\subsection{Experimental Setup}
\textbf{Datasets.} ~~In this work, we evaluate our proposed method on two benchmark datasets to comprehensively assess its performance and effectiveness in detecting and mitigating hallucinations in video-language models.

The first dataset is the \textit{ActivityNet Captions} \cite{krishna2017dense} dataset, which we primarily use for detecting hallucinations. It is built on ActivityNet v1.3 and contains 20,000 untrimmed YouTube videos with 100,000 human-written caption annotations. The average video length is 120 seconds, and most videos contain 3 or more annotated events, with each event having a start time, end time, and a corresponding caption. The average caption length is 13.5 words, making it suitable for evaluating the semantic consistency between generated captions and reference contexts.

The second dataset is \textit{ActivityNet-QA}, which is used to assess the overall accuracy and question-answering capability of our model in a zero-shot evaluation setting. Derived from the ActivityNet \cite{ActivityNet} dataset, it comprises 5,800 untrimmed web videos, each associated with ten human-annotated question-answer (QA) pairs, totaling 58,000 QA pairs. The dataset is partitioned into training, validation, and test sets, containing 32,000 QA pairs across 3,200 videos, 18,000 QA pairs across 1,800 videos, and 8,000 QA pairs across 800 videos, respectively. The videos cover a diverse range of activities, providing a robust foundation for assessing our models' performance on complex video content.

\textbf{Baselines.} ~~For baseline comparisons, we perform two separate sets of evaluations. The first set compares the model's faithfulness score before and after the application of our hallucination mitigation techniques. This comparison demonstrates the effectiveness of our retrieval-augmented generation (RAG) module in reducing hallucinations in the generated captions. The second set of comparisons evaluates the model's overall accuracy against several state-of-the-art methods. Specifically, our model, ResNetVLLM + RAG, is compared against the original ResNetVLLM model without RAG, as well as BT-Adapter, Valley, Video-ChatGPT, Video LLaMA, LLaMA Adapter v2, Video Chat, and FrozenBiLM.

\textbf{Implementation Details.} ~~We now present the implementation details of our hallucination detection and mitigation protocols.

The \textit{hallucination detection} experiments are implemented using PyTorch and the Hugging Face Transformers library. We extend the Lynx hallucination detection model by adapting its input pipeline to accept video-generated captions and corresponding ground-truth captions. The model is fine-tuned on the ActivityNet Captions dataset, where each training instance consists of a pair of generated and reference captions. The input text pairs are tokenized using a pre-trained T5 tokenizer, and the model is trained to predict a scalar faithfulness score using a regression head on top of the encoder’s final hidden state. During fine-tuning, we use a batch size of 32 and a learning rate of \(2 \times 10^{-5}\), optimized using the AdamW optimizer with a linear learning rate scheduler and warm-up steps. Training is conducted for 5 epochs with early stopping based on validation loss. All experiments are run on a workstation equipped with four NVIDIA A100 GPUs (80 GB each), enabling parallelized training and evaluation.

For the \textit{hallucination mitigation} experiments are implemented by extending the ResNetVLLM framework \cite{khalil2024resnetvllm}. The backbone consists of a ResNet-50 \cite{he2015deepresiduallearningimage} visual encoder and the LLaVA \cite{liu2023visual} transformer-based language model, jointly trained on the Video-ChatGPT-100K instruction dataset \cite{maaz2023videochatgpt} dataset. ResNet processes 100 sampled frames per video, with each frame resized to 224×224 and passed through a projection layer. Extracted features, along with video metadata, are used to construct an ad-hoc knowledge base at inference time. ~The language model, initialized from pretrained LLaVA weights, is fine-tuned jointly with the visual encoder. Training is performed in two phases: a warm-up stage using SGD for 150 epochs (learning rate 0.01, weight decay 0.0001), and a joint-training phase using AdamW for 50 epochs (learning rate 0.00015, weight decay 0.05). Hyperparameters are optimized using Bayesian Optimization.

The Post-Hoc RAG component is implemented following the RARR framework \cite{gao2023rarr}, where the model performs retrieval over an ad-hoc knowledge base and uses the retrieved evidence to assess and revise the generated caption. Evaluation is performed using a faithfulness score computed by a modified Lynx model. All experiments are implemented in PyTorch with mixed-precision training enabled. Training was conducted on 2×Tesla V100 GPUs and took approximately 4 hours. Key libraries include Hugging Face Transformers, TorchVision, and OpenCV.

\subsection{Evaluation Metrics}
\textbf{Faithfulness Score for Hallucination Detection.} ~~The \textit{faithfulness score} \cite{ravi2024lynxopensourcehallucination} is used to evaluate the semantic alignment between the generated and ground-truth captions. It measures how accurately the generated caption reflects the content described by the reference context. The score is computed by averaging the cosine similarity between the embeddings of the generated and ground-truth captions across all video-text pairs. The formula for the faithfulness score \cite{ravi2024lynxopensourcehallucination} is defined as:

\begin{equation}
    \label{equ:faithfulnness_score}
    \textrm{Faithfulness} = \frac{1}{N} \sum_{i=1}^{N} \cos(\text{emb}_{\text{generated}}^i, \text{emb}_{\text{ground truth}}^i)
\end{equation}

Where:
\begin{itemize}
    \item \( N \) is the number of video-text pairs,
    \item \( \text{emb}_{\text{generated}}^i \) and \( \text{emb}_{\text{ground truth}}^i \) represent the embeddings of the generated and ground-truth captions, respectively, and
    \item \( \cos \) denotes the cosine similarity function.
\end{itemize}
A higher faithfulness score (above 50\%) indicates greater consistency between the generated and reference captions, suggesting the absence of hallucination. In contrast, a lower score (below 50\%) indicates a substantial deviation from the context of the truth of the ground, which indicates the presence of hallucination.

\textbf{Accuracy.} We evaluated the performance of ResNetVLLM-2 using \textit{accuracy}, a widely used metric for classification tasks \cite{xu2017video}. Accuracy measures the proportion of correctly predicted answers in the total number of test samples. Let \( N \) denote the size of the test set, where each question \( \mathbf{q}_i \in Q \) has a corresponding ground-truth answer \( \mathbf{y}_i \in Y \). The predicted answer of the model for question \( \mathbf{q}_i \) is represented by \( \mathbf{a}_i \). The accuracy is calculated using the following formula \cite{yu2019activitynet}:
\begin{equation}
    \textrm{Accuracy} = \frac{1}{N} \sum\limits_{i=1}^{N} \mathbf{1}[\mathbf{a}_i = \mathbf{y}_i]
\end{equation}

Where:
\begin{itemize}
    \item \( \mathbf{1}[\cdot] \) is an indicator function that returns \( 1 \) if the predicted answer \( \mathbf{a}_i \) matches the ground-truth answer \( \mathbf{y}_i \), and \( 0 \) otherwise.
\end{itemize}

\subsection{Results}
\textbf{ResNetVLLM Faithfulness Score.} ~~Table \ref{tab:resnetvllm-faithfulness} shows a substantial improvement in the faithfulness score of ResNetVLLM on the ActivityNet Captions dataset, increasing from 34.2\% before mitigation to 97.9\% after applying the hallucination mitigation techniques, indicating a considerable reduction in hallucination.
\begin{table}
    \centering
    \begin{tabular}{lcc}
        \toprule
        \textbf{Model} & \textbf{Before Mitigation} & \textbf{After Mitigation} \\
        \midrule
        ResNetVLLM & 34.2\% & 92.7\% \\
        \bottomrule
    \end{tabular}
    \caption{Faithfulness Score of ResNetVLLM on ActivityNet Captions before and after hallucination mitigation.}
    \label{tab:resnetvllm-faithfulness}
\end{table}

\textbf{Accuracy (Zero-Shot Question-Answer Evaluation).} ~~To evaluate our model, we compared its accuracy against several notable models, such as FrozenBiLM \cite{yang2022zero}, and the generative video models Video Chat \cite{li2023videochat}, LLaMA Adapter \cite{gao2023llamaadapter}, Video LLaMA \cite{zhang2023videollama}, and Video-ChatGPT \cite{maaz2023videochatgpt}. After introducing hallucination mitigation techniques in ResNetVLLM, we observed a notable improvement in its performance, with the accuracy increasing from 54.8\% to 65.3\% on the ActivityNet-QA benchmark. This improvement highlights the effectiveness of the proposed methods in enhancing the model's ability to generate more accurate and contextually relevant captions. The results are displayed in Table \ref{tab:zero-shot-question-answer-evaluation}.
\begin{table}[!t]
    \centering
    \setlength{\tabcolsep}{6pt}
    \renewcommand{\arraystretch}{1}
    \caption{Comparison with recent state-of-the-art methods on the \textbf{The Zero-Shot Question-Answer Evaluation} component of the \textbf{video understanding benchmark} \cite{maaz2023videochatgpt} on \textbf{ActivityNet-QA} \cite{yu2019activitynet}.}
    \label{tab:zero-shot-question-answer-evaluation}
    \begin{tabular}{l c}
        \hline
        \textbf{Model} & \textbf{Activity Net-QA} \\
        \hline
        FrozenBiLM \cite{yang2022zero} & 24.7 \\
        Video Chat \cite{li2023videochat} & 26.5 \\
        LLaMA Adapter v2 \cite{gao2023llamaadapter} & 34.2 \\
        Video LLaMA \cite{zhang2023videollama} & 12.4 \\
        Video-ChatGPT \cite{maaz2023videochatgpt} & 35.2 \\
        Valley \cite{luo2023valley} & 45.1 \\
        BT-Adapter \cite{liu2023one} & 45.7 \\
        ResNetVLLM \cite{khalil2024resnetvllm} & 54.8 \\
        \hline
        \textbf{ResNetVLLM-2} & \textbf{68.3} \\
        \hline
    \end{tabular}
\end{table}

\section{Conclusion}
In this paper, we presented ResNetVLLM-2, an improved version of the ResNetVLLM model designed to address multi-modal hallucinations in video-language tasks. By incorporating a faithfulness detection strategy and a retrieval-augmented generation (RAG) module with a dynamically built ad-hoc knowledge base, we substantially enhance the model's factual consistency. Our evaluation on the ActivityNet-QA benchmark demonstrates a 10.5\% accuracy improvement, highlighting the effectiveness of our hallucination mitigation protocol. The proposed approach offers a scalable and generalizable solution for improving the reliability of video-language models, making them more suitable for real-world applications such as video analysis, autonomous systems, and multimedia retrieval. Future work will focus on extending the RAG framework to support multi-turn video dialogues and exploring fine-grained temporal alignment to further reduce hallucinations.

\bibliographystyle{IEEEtran}
\bibliography{IEEEabrv,paper}

\onecolumn 
\appendices

\newpage
\section{Example of Hallucination in Vision}
\label{app:A-Appendix}
\renewcommand{\thefigure}{A.\arabic{figure}}
\begin{figure}[H] 
    \centering
    \begin{minipage}{0.8\textwidth} 
        \centering
        \includegraphics[width=\linewidth]{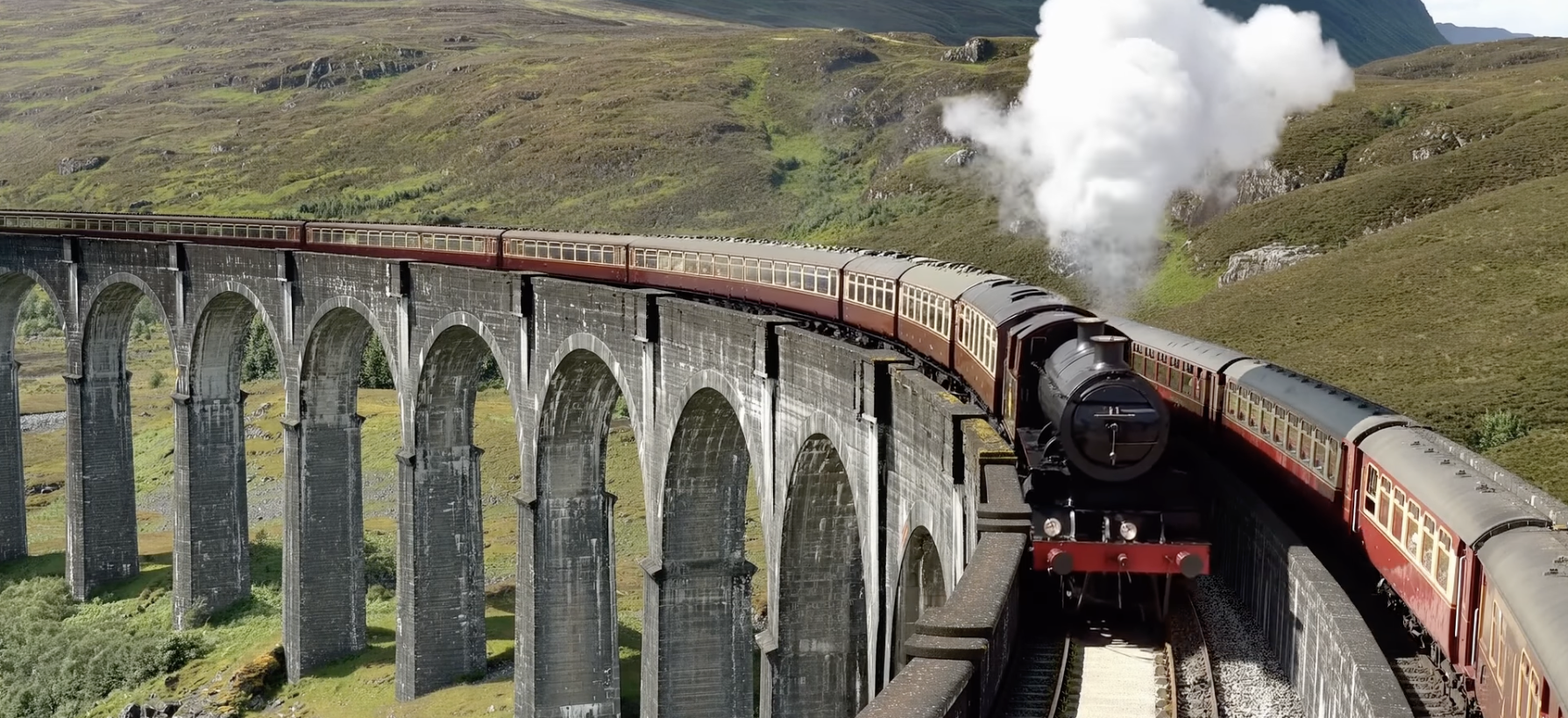}
        \caption{A Sora-generated video of the \textbf{Glenfinnan Viaduct} in Scotland, incorrectly showing a \textbf{second track} whereas the real viaduct has \textbf{only one}, a \textbf{second chimney} on its interpretation of the train \textbf{The Jacobite}, and some carriages much longer than others. \cite{OpenAI2024}}
        \label{fig:example-vllm-hallucination-Glenfinnan}
    \end{minipage}
    \vfill
    \vspace{0.5cm}
    \begin{minipage}{0.55\textwidth} 
        \centering
        \includegraphics[width=\linewidth]{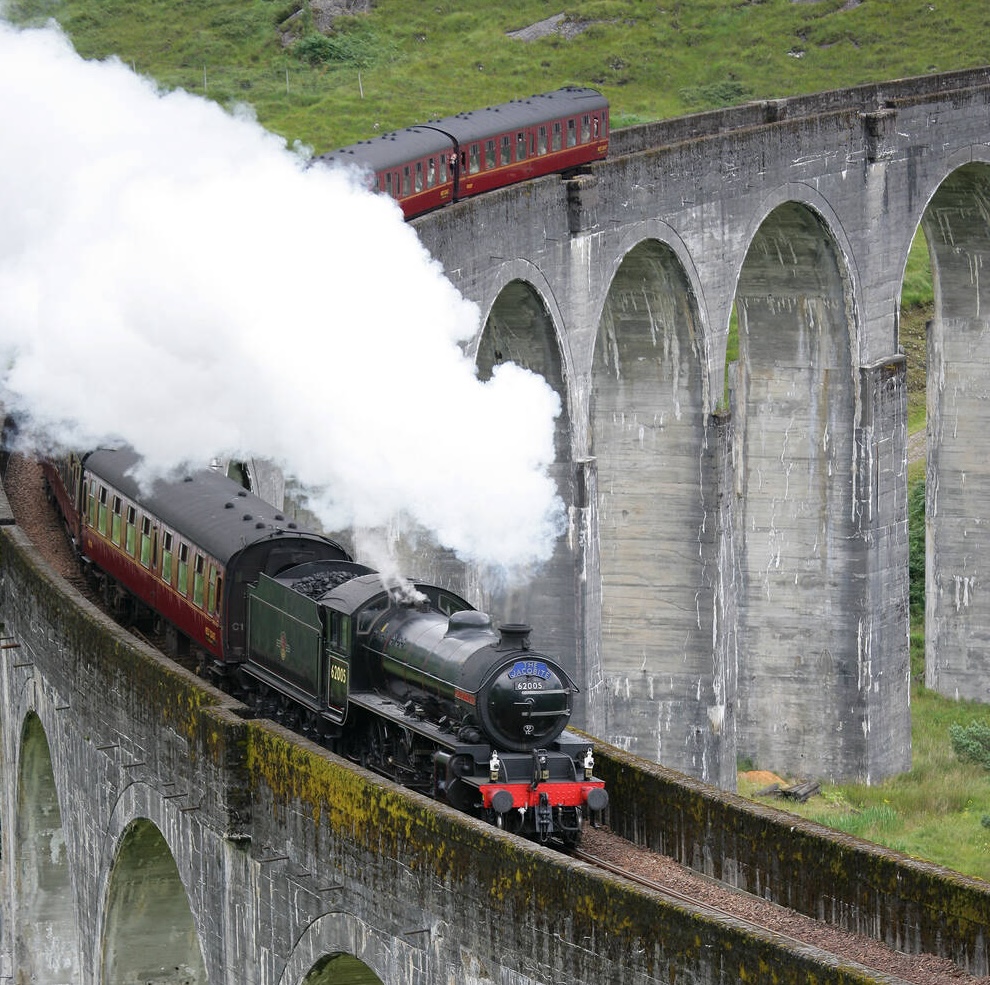}
        \caption{The \textbf{Glenfinnan Viaduct} in Scotland \cite{glenfinnan_viaduct}}
        \label{fig:real_glenfinnan_viaduct}
    \end{minipage}
\end{figure}

\newpage
\section{Hallucination vs. Fake vs. Error vs. Incorrectness}
\label{app:B-Appendix}
\renewcommand{\thefigure}{B.\arabic{figure}}
In the context of AI-generated content, the terms hallucination, fake, error, and incorrectness are often used interchangeably, but they carry distinct meanings depending on the cause, intent, and context. This section clarifies the differences between these terms.

\subsection{Fake} \textbf{Definition:} Intentionally false information designed to deceive or manipulate.
\textbf{Example:} A fabricated news article spreading misinformation about a political event.
\textbf{Context:} Associated with disinformation, propaganda, or media manipulation.
\textbf{Key Trait:} The falsehood is intentional and aimed at misleading the audience.

\subsection{Error} \textbf{Definition:} A mistake made by the AI, such as factual inaccuracies, logical inconsistencies, or language errors.
\textbf{Example:} An AI incorrectly solving a simple arithmetic problem, e.g., stating \(2 + 2 = 5\).
\textbf{Context:} Typically arises from flawed algorithms, poor training data, or limited model accuracy.
\textbf{Key Trait:} Errors are usually identifiable and correctable upon review.

\subsection{Incorrectness} \textbf{Definition:} A general term for any kind of inaccuracy, including factual, logical, or grammatical mistakes.
\textbf{Example:} An AI providing the wrong capital city for a country.
\textbf{Context:} Broadly covers all types of inaccuracies without specifying the cause or intent.
\textbf{Key Trait:} It does not necessarily imply intent or a specific source of the mistake.

\subsection{Comparison Summary} Table~\ref{tab:comparison} summarizes the key differences between these terms.

\begin{table}[h] 
    \centering
    \caption{Comparison of Hallucination, Fake, Error, and Incorrectness.}
    \label{tab:comparison}
    \begin{tabular}{lccc}
        \toprule
        \textbf{Term} & \textbf{Cause} & \textbf{Intentional?} & \textbf{Example} \\
        \midrule Hallucination & AI generates false info & No & AI says "Blue apples grow in Antarctica." \\
        Fake & False info created to deceive & Yes & Fake news article spreading false rumors \\
        Error & Mistake or miscalculation & No & AI says \(2 + 2 = 5\) \\
        Incorrectness & General inaccuracy & Not always & AI says the Eiffel Tower is in Berlin \\ 
        \bottomrule
    \end{tabular}
\end{table}

\newpage
\section{Related Work Appendix}
\renewcommand{\thefigure}{C.\arabic{figure}}
\begin{figure}[ht]
    \centering
    \includegraphics[width=0.75\linewidth]{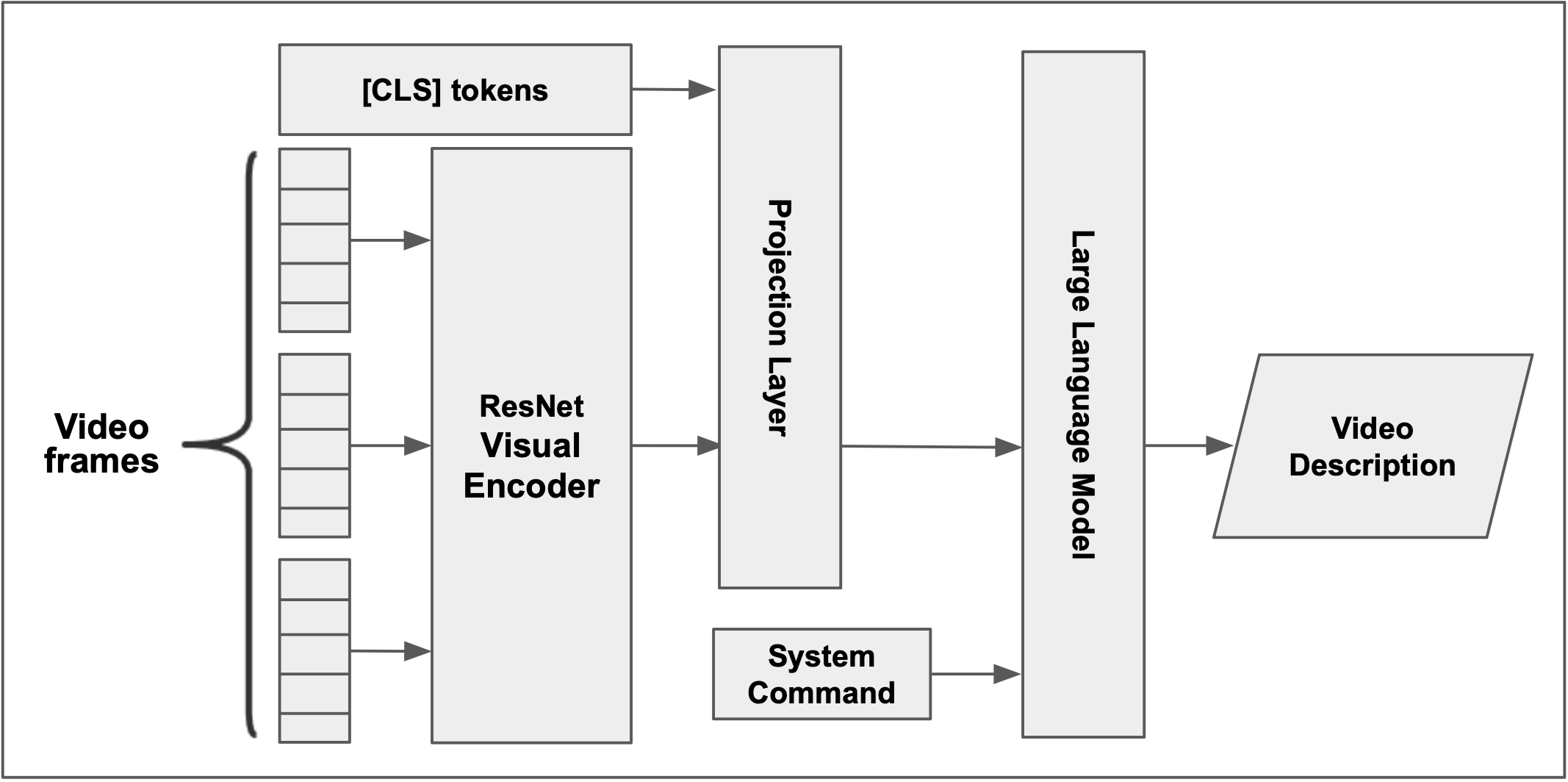}
    \caption{Overall architecture of the ResNetVLLM model. \cite{khalil2024resnetvllm}}
    \label{fig:resnetvllm-architecture}
\end{figure}

\begin{figure}[h]
    \centering
    \includegraphics[width=1\linewidth]{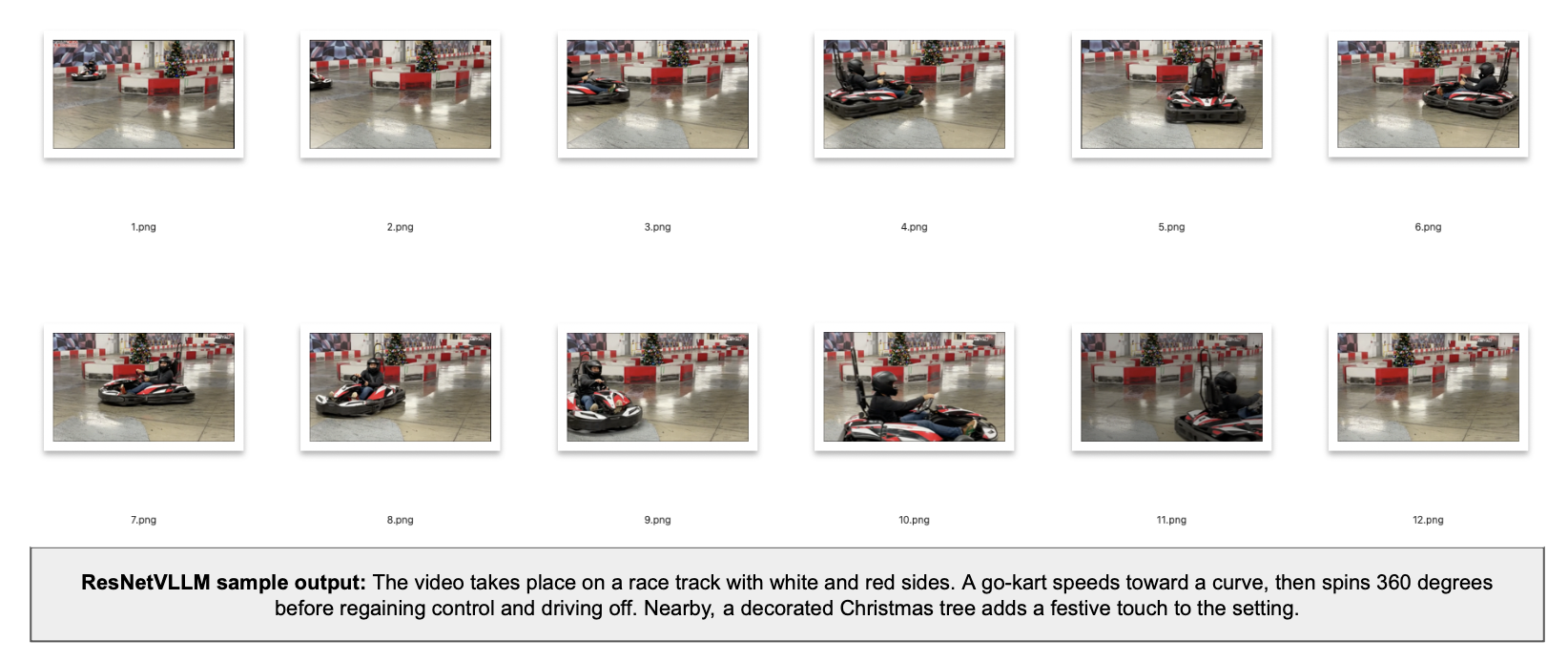}
    \caption{Sample output of ResNetVLLM. \cite{khalil2024resnetvllm}}
    \label{fig:resnetvllm-output}
\end{figure}

\begin{figure}[h]
    \centering
    \includegraphics[width=0.6\linewidth]{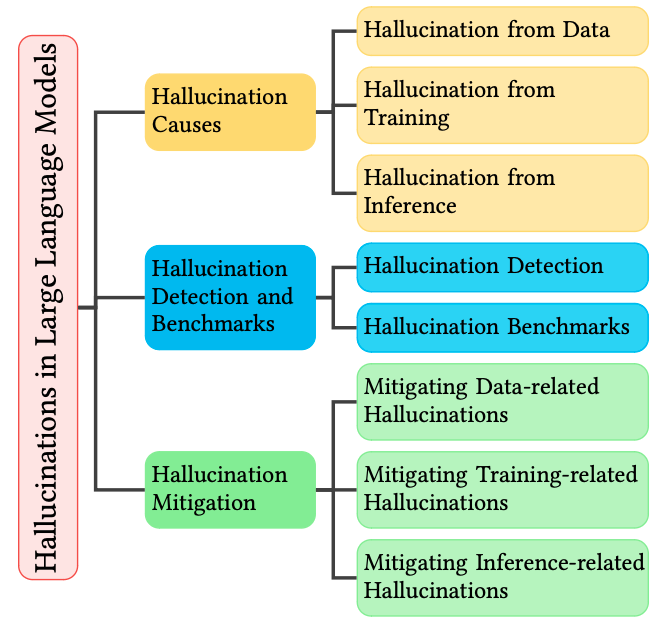}
    \caption{The main content flow of the survey on hallucination in large language models by Huang et al. \cite{Huang_2025}}
    \label{fig:hallucination-survey}
\end{figure}

\begin{table}[!t]
    \centering
    \setlength{\tabcolsep}{6pt}
    \renewcommand{\arraystretch}{1.2}
    \caption{Comparison with recent state-of-the-art methods on the \textbf{Zero-Shot Question-Answer Evaluation} component of the \textbf{video understanding benchmark} \cite{maaz2023videochatgpt} on \textbf{ActivityNet-QA} \cite{yu2019activitynet}.}
    \label{tab:zero-shot-evaluation}
    \begin{tabular}{l c}
        \hline
        \textbf{Model} & \textbf{ActivityNet-QA} \\
        \hline
        FrozenBiLM \cite{yang2022zero} & 24.7 \\
        Video Chat \cite{li2023videochat} & 26.5 \\
        LLaMA Adapter v2 \cite{gao2023llamaadapter} & 34.2 \\
        Video LLaMA \cite{zhang2023videollama} & 12.4 \\
        Video-ChatGPT \cite{maaz2023videochatgpt} & 35.2 \\
        Valley \cite{luo2023valley} & 45.1 \\
        BT-Adapter \cite{liu2023one} & 45.7 \\
        \hline
        \textbf{ResNetVLLM \cite{khalil2024resnetvllm}} & \textbf{54.8} \\
        \hline
    \end{tabular}
\end{table}

\end{document}